\documentclass[runningheads]{llncs}
\usepackage{graphicx}
\usepackage{amsmath,amssymb}
\usepackage{color}
\usepackage{soul}
\usepackage[width=122mm,left=12mm,paperwidth=146mm,height=193mm,top=12mm,paperheight=217mm]{geometry}
\usepackage[pagebackref=true,breaklinks=true,colorlinks=false,bookmarks=false]{hyperref}
\usepackage{booktabs}
\usepackage{xspace}
\begin{document}
\pagestyle{headings}
\mainmatter
\def\ECCV16SubNumber{000}

\title{Hollywood in Homes: Crowdsourcing Data Collection for Activity Understanding}

\makeatletter
\DeclareRobustCommand\onedot{\futurelet\@let@token\@onedot}
\def\@onedot{\ifx\@let@token.\else.\null\fi\xspace}
\def\eg{\emph{e.g}\onedot} \def\Eg{\emph{E.g}\onedot}
\def\ie{\emph{i.e}\onedot} \def\Ie{\emph{I.e}\onedot}
\def\cf{\emph{c.f}\onedot} \def\Cf{\emph{C.f}\onedot}
\def\etc{\emph{etc}\onedot} \def\vs{\emph{vs}\onedot}
\def\wrt{w.r.t\onedot} \def\dof{d.o.f\onedot}
\def\etal{\emph{et al}\onedot}
\makeatother

\titlerunning{Hollywood in Homes: Crowdsourcing Data Collection}

\authorrunning{Sigurdsson, Varol, Wang, Farhadi, Laptev, Gupta}

\newcommand{\repeatthanks}{\textsuperscript{\thefootnote}}
\newcommand{\authSpace}{,\ \ }
\author{
Gunnar A. Sigurdsson$^{1}$\authSpace 
G\"ul Varol$^{2}$\authSpace
Xiaolong Wang$^{1}$\authSpace \\
Ali Farhadi$^{3,4}$\authSpace  
Ivan Laptev$^{2}$\authSpace and  
Abhinav Gupta$^{1,4}$
}
\newcommand{\ispace}{\ \ \ \ }
\institute{
 $^{1}$Carnegie Mellon University \ispace
  $^{2}$Inria \ispace
  $^{3}$University of Washington \\
 $^{4}$The Allen Institute for AI \\ \vspace{0.3cm}
 \url{http://allenai.org/plato/charades/}
 }

\maketitle

\begin{abstract}
Computer vision has a great potential to help our daily lives by searching for lost keys, watering flowers or reminding us to take a pill. To succeed with such tasks, computer vision methods need to be trained from real and diverse examples of our daily dynamic scenes. While most of such scenes are not particularly exciting, they typically do not appear on YouTube, in movies or TV broadcasts. So how do we collect sufficiently many diverse but {\em boring} samples representing our lives? We propose a novel Hollywood in Homes approach to collect such data. Instead of shooting videos in the lab, we ensure diversity by distributing and crowdsourcing the whole process of video creation from script writing to video recording and annotation. Following this procedure we collect a new dataset, \textit{Charades}, with hundreds of people recording videos in their own homes, acting out casual everyday activities. The dataset is composed of 9,848 annotated videos with an average length of 30 seconds, showing activities of 267 people from three continents, and over $15\%$ of the videos have more than one person. Each video is annotated by multiple free-text descriptions, action labels, action intervals and classes of interacted objects. In total, Charades provides 27,847 video descriptions, 66,500 temporally localized intervals for 157 action classes and 41,104 labels for 46 object classes. Using this rich data, we evaluate and provide baseline results for several tasks including action recognition and automatic description generation. We believe that the realism, diversity, and casual nature of this dataset will present unique challenges and new opportunities for computer vision community.
\end{abstract}

\section{Introduction}
Large scale visual learning fueled by huge datasets has changed the computer vision landscape~\cite{deng2009imagenet,zhou2014learning}. Given the source of this data, it's not surprising that most of our current success is biased towards static scenes and objects in Internet images. As we move forward into the era of AI and robotics, however, new questions arise. How do we learn about different states of objects (\eg, cut vs. whole)? How do common activities affect changes of object states? In fact, it is not even yet clear if the success of the Internet pre-trained recognition models will transfer to real-world settings where robots equipped with our computer vision models should operate.

Shifting the bias from Internet images to real scenes will most likely require
collection of new large-scale datasets representing activities of  
our boring everyday life: getting up, getting dressed, putting groceries in fridge, cutting vegetables and so on. Such datasets will allow us to develop new representations and to learn models with the right biases. But more importantly, such datasets representing people interacting with objects and performing natural action sequences in typical environments will finally allow us to learn common sense and contextual knowledge necessary for high-level reasoning and modeling.

\begin{figure}[t]
    \centering
    \includegraphics[width=0.8\linewidth]{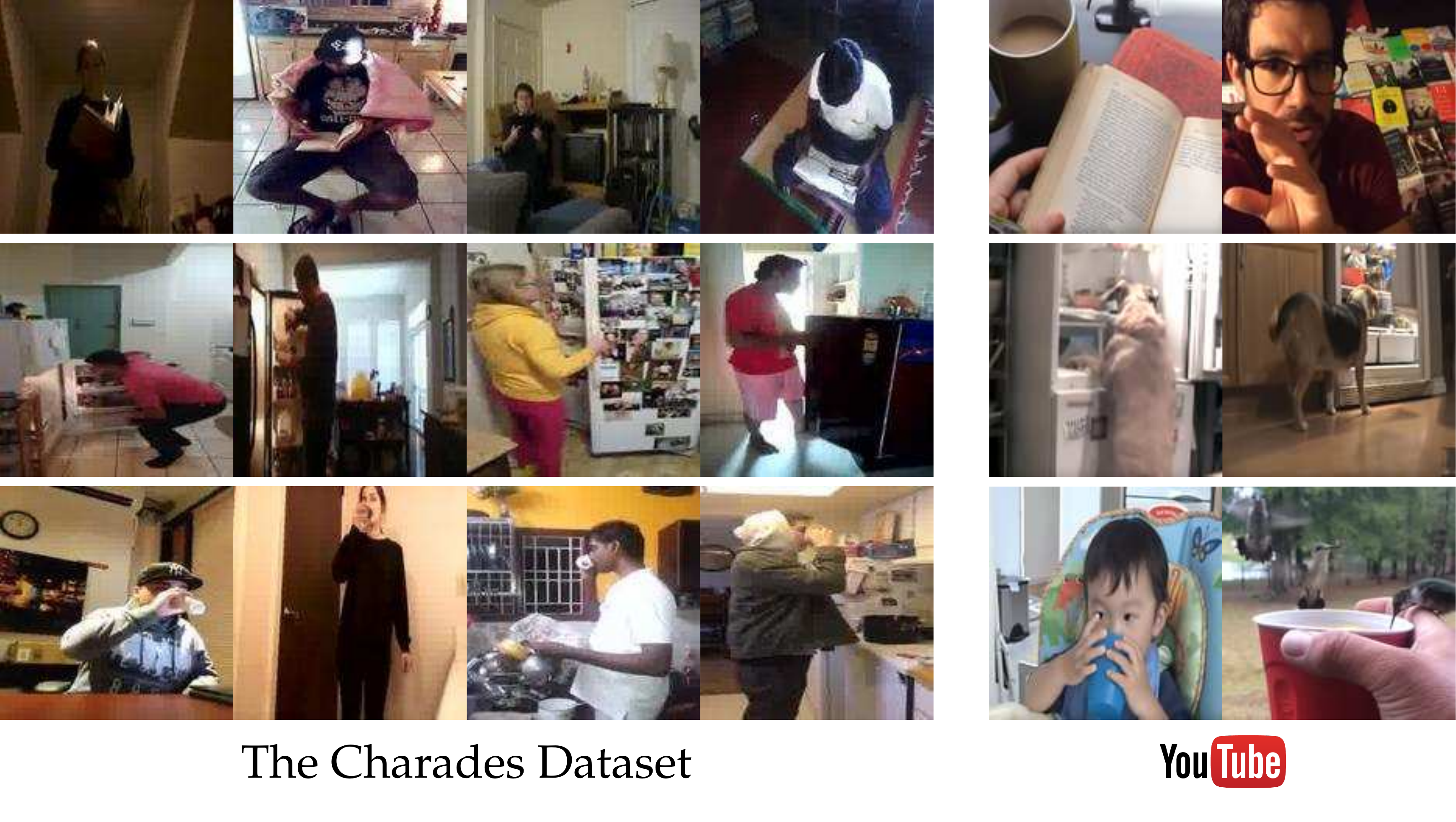}
    \caption{Comparison of actions in the Charades dataset and on YouTube: \textit{Reading a book}, \textit{Opening a refrigerator}, \textit{Drinking from a cup}. YouTube returns entertaining and often atypical videos, while \textit{Charades} contains typical everyday videos.}
    \label{fig:teaser}
\end{figure}

But how do we find these boring videos of our daily lives? If we search common activities such as ``drinking from a cup'', ``riding a bike'' on video sharing websites such as YouTube, we observe a highly-biased sample of results (see Figure~\ref{fig:teaser}). These results are biased towards entertainment---boring videos have no viewership and hence no reason to be uploaded on YouTube!

In this paper, we propose a novel {\bf Hollywood in Homes} approach to collect a large-scale dataset of boring videos of daily activities. Standard approaches in the past have used videos downloaded from the Internet~\cite{caba2015activitynet,liu2009recognizing,THUMOS15,karpathy2014large,kuehne2011hmdb,UCF101} gathered from movies~\cite{laptev2008learning,rodriguez2008action,rohrbach15cvpr} or recorded in controlled environments~\cite{schuldt2004recognizing,ActionsAsSpaceTimeShapes_pami07,rohrbach2012database,oh2011large,Kuehne12,rohrbach2014coherent}. Instead, as the name suggests: we take the Hollywood filming process to the homes of hundreds of people on Amazon Mechanical Turk (AMT). AMT workers follow the three steps of filming process: (1) script generation; (2) video direction and acting based on scripts; and (3) video verification to create one of the largest and most diverse video dataset of daily activities. 

There are threefold advantages of using the {\bf Hollywood in Homes} approach for dataset collection:  (a) Unlike datasets shot in controlled environments (\eg, MPII~\cite{rohrbach2012database}), crowdsourcing brings in diversity which is essential for generalization. In fact, our approach even allows the same script to be enacted by multiple people; (b) crowdsourcing the script writing enhances the coverage in terms of scenarios and reduces the bias introduced by generating scripts in labs; and (c) most importantly, unlike for web videos, this approach allows us to control the composition and the length of video scenes by proposing the vocabulary of scenes, objects and actions during script generation.

\noindent {\bf The Charades v1.0 Dataset}

\noindent \emph{Charades} is our large-scale dataset with a focus on common household activities collected using the Hollywood in Homes approach. The name comes from of a popular American word guessing game where one player acts out a phrase and the other players guess what phrase it is. In a similar spirit, we recruited hundreds of people from Amazon Mechanical Turk to act out a paragraph that we presented to them.
The workers additionally provide action classification, localization, and video description annotations. The first publicly released version of our {\em Charades} dataset will contain $9{,}848$ videos of daily activities $30.1$ seconds long on average ($7{,}985$ training and $1{,}863$ test). The dataset is collected in 15 types of indoor scenes, involves interactions with $46$ object classes and has a vocabulary of $30$ verbs leading to $157$ action classes. It has $66{,}500$ temporally localized actions, $12.8$ seconds long on average, recorded by 267 people in three continents. We believe this dataset will provide a crucial stepping stone in developing action representations, learning object states, human object interactions, modeling context, object detection in videos, video captioning and many more. The dataset will be publicly available at \url{http://allenai.org/plato/charades/}.

\noindent \textbf{Contributions} The contributions of our work are three-fold: (1) We introduce the Hollywood in Homes approach to data collection, (2) we collect and release the first crowdsourced large-scale dataset of boring household activities, and (3) we provide extensive baseline evaluations.

The KTH action dataset~\cite{schuldt2004recognizing} paved the way for algorithms that recognized human actions. However, the dataset was limited in terms of number of categories and enacted in the same background. In order to scale up the learning and the complexity of the data, recent approaches have instead tried collecting video datasets by downloading videos from Internet. Therefore, datasets such as
UCF101~\cite{UCF101}, Sports1M~\cite{karpathy2014large} and others~\cite{kuehne2011hmdb,liu2009recognizing,THUMOS15} appeared and presented more challenges including background clutter, and scale. However, since it is impossible to find boring daily activities on Internet, the vocabulary of actions became biased towards more sports-like actions which are easy to find and download.

There have been several efforts in order to remove the bias towards sporting actions. One such commendable effort is to use movies as the source of data~\cite{marszalek09,ferrari20092d}. Recent papers have also used movies to focus on the video description problem leading to several datasets such as
MSVD\cite{chen2011collecting}, M-VAD~\cite{torabi2015using}, and MPII-MD~\cite{rohrbach15cvpr}. Movies however are still exciting (and a source of entertainment) and do not capture the scenes, objects or actions of daily living. Other efforts have been to collect in-house datasets for capturing human-object interactions~\cite{gupta2007objects} or human-human interactions~\cite{ryoo2009spatio}. Some relevant big-scale efforts in this direction include MPII Cooking~\cite{rohrbach2012database}, TUM Breakfast~\cite{Kuehne12}, and the TACoS Multi-Level~\cite{rohrbach2014coherent} datasets. These datasets focus on a narrow domain by collecting the data in-house with a fixed background, and therefore focus back on the activities themselves. This allows for careful control of the data distribution, but has limitations in terms of generalizability, and scalability. In contrast, PhotoCity~\cite{tuite2011photocity} used the crowd to take pictures of landmarks, suggesting that the same could be done for other content at scale.

Another relevant effort in collection of data corresponding to daily activities and objects is in the domain of ego-centric cameras. For example, the Activities of Daily Living dataset~\cite{pirsiavash2012detecting} recorded 20 people performing unscripted, everyday activities in their homes in first person, and another extended that idea to animals~\cite{yumi2014first}. These datasets provide a challenging task but fail to provide diversity which is crucial for generalizability. It should however be noted that these kinds of datasets could be crowdsourced similarly to our work.

The most related dataset is the recently released ActivityNet dataset~\cite{caba2015activitynet}. It includes actions of daily living downloaded from YouTube. We believe the ActivityNet effort is complementary to ours since their dataset is uncontrolled, slightly biased towards non-boring actions and biased in the way the videos are professionally edited. On the other hand, our approach focuses more on action sequences (generated from scripts) involving interactions with objects. Our dataset, while diverse, is controlled in terms of vocabulary of objects and actions being used to generate scripts. In terms of the approach, Hollywood in Homes is also related to \cite{zitnick2013bringing}. However, \cite{zitnick2013bringing} only generates synthetic data. A comparison with other video datasets is presented in Table~\ref{tbl:comparison}. To the best of our knowledge, our approach is the first to demonstrate that workers can be used to collect a vision dataset by filming themselves at such a large scale.

\begin{table}[t]
\centering
\caption{Comparison of Charades with other video datasets.}
\label{tbl:comparison}
\resizebox{\textwidth}{!}{%
\begin{tabular}{@{\hspace{.1in}}llllllll@{}}
\toprule
 & \multicolumn{1}{c}{\begin{tabular}[c]{@{}c@{}}Actions\\ per video\end{tabular}} & \multicolumn{1}{c}{\begin{tabular}[c]{@{}c@{}}Classes\\ \end{tabular}} & \multicolumn{1}{c}{\begin{tabular}[c]{@{}c@{}}Labelled \\ instances\end{tabular}} & \multicolumn{1}{c}{\begin{tabular}[c]{@{}c@{}}Total \\ videos\end{tabular}} & \multicolumn{1}{c}{\begin{tabular}[c]{@{}c@{}}Origin\\ \end{tabular}} & \multicolumn{1}{c}{\begin{tabular}[c]{@{}c@{}}Type\\ \end{tabular}} & \multicolumn{1}{c}{\begin{tabular}[c]{@{}c@{}}Temporal \\ localization\end{tabular}} \\ \midrule
\rule{0pt}{2.3ex}Charades v1.0 & 6.8 & 157 & 67K & 10K & 267 Homes & Daily Activities & Yes \\ \midrule
\rule{0pt}{3ex}ActivityNet~\cite{caba2015activitynet} & 1.4 & 203 & 39K & 28K & YouTube & Human Activities & Yes \\
UCF101~\cite{UCF101} & 1 & 101 & 13K & 13K & YouTube & Sports & No \\
HMDB51~\cite{kuehne2011hmdb} & 1 & 51 & 7K & 7K & YouTube/Movies & Movies & No \\
THUMOS'15~\cite{THUMOS15} & 1-2 & 101 & 21K+ & 24K & YouTube & Sports & Yes \\
Sports 1M~\cite{karpathy2014large} & 1 & 487 & 1.1M & 1.1M & YouTube & Sports & No \\
MPII-Cooking~\cite{rohrbach2012database} & 46 & 78 & 13K & 273 & 30 In-house actors & Cooking & Yes \\
ADL~\cite{pirsiavash2012detecting} & 22 & 32 & 436 & 20 & 20 Volunteers & Ego-centric & Yes \\
MPII-MD~\cite{rohrbach15cvpr} & Captions & Captions & 68K & 94 & Movies & Movies & No \\ \bottomrule
\end{tabular}%
}
\end{table}

\section{Hollywood in Homes}
\label{sec:charades}

We now describe the approach and the process involved in a large-scale video collection effort via AMT. Similar to filming, we have a three-step process for generating a video. The first step is generating the script of the indoor video. The key here is to allow workers to generate diverse scripts yet ensure that we have enough data for each category. The second step in the process is to use the script and ask workers to record a video of that sentence being acted out. In the final step, we ask the workers to verify if the recorded video corresponds to script, followed by an annotation procedure.

\subsection{Generating Scripts} In this work we focus on indoor scenes, hence, we group together rooms in residential homes (\textit{Living Room}, \textit{Home Office}, etc.). We found 15 types of rooms to cover most of typical homes, these rooms form the scenes in the dataset.
In order to generate the \textit{scripts} (a text given to workers to act out in a video), we use a vocabulary of objects and actions to guide the process. To understand what objects and actions to include in this vocabulary, we analyzed 549 movie scripts from popular movies in the past few decades.
Using both term-frequency (TF) and TF-IDF~\cite{salton1983mcgill} we analyzed which nouns and verbs occur in those rooms in these movies. From those we curated a list of 40 objects and 30 actions to be used as seeds for script generation, where objects and actions were chosen to be generic for different scenes.

\begin{figure}[t]
    \centering
    \includegraphics[width=1.0\linewidth]{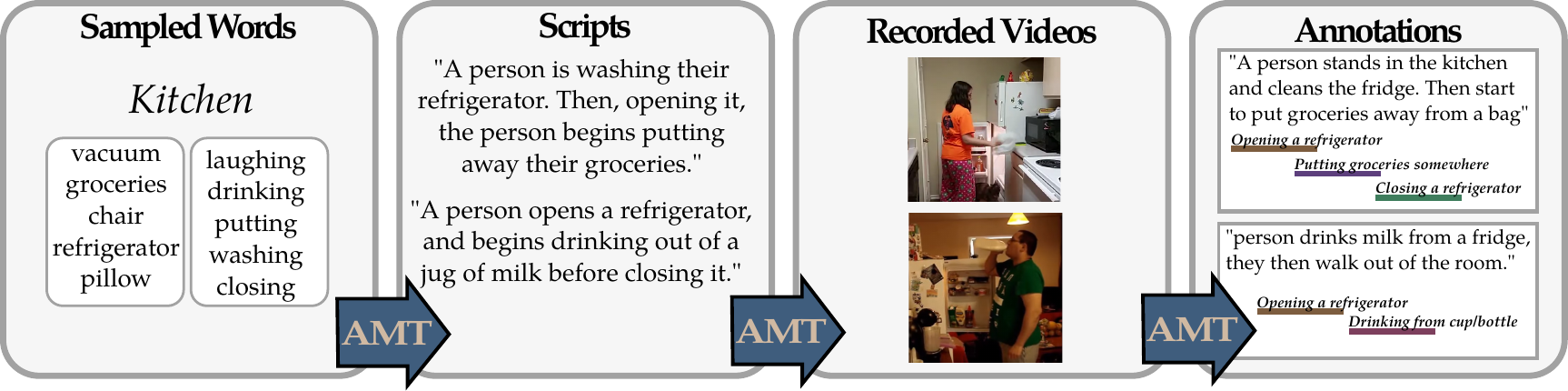}
    \caption{An overview of the three Amazon Mechanical Turk (AMT) crowdsourcing stages in the \emph{Hollywood in Homes} approach.}
    \label{fig:mturkpipeline}
\end{figure}

To harness the creativity of people, and understand their bias towards activities, we crowdsourced the script generation as follows. In the AMT interface, a single scene, 5 randomly selected objects, and 5 randomly selected actions were presented to workers. Workers were asked to use 
two objects and two actions to compose a short paragraph about activities of one or two people performing realistic and commonplace activities in their home. We found this to be a good compromise between controlling what kind of words were used and allowing the users to impose their own human bias on the generation. Some examples of generated scripts are shown in Figure~\ref{fig:mturkpipeline}. (see the website for more examples). 
The distribution of the words in the dataset is presented in Figure~\ref{fig:mega}.

\subsection{Generating Videos}
Once we have scripts, our next step is to collect videos. To maximize the diversity of scenes, objects, clothing and behaviour of people, we ask the workers themselves to record the 30 second videos by following collected scripts. 

AMT is a place where people commonly do quick tasks in the convenience of their homes or during downtime at their work. AMT has been used for annotation and editing but can we do content creation via AMT? During a pilot study we asked workers to record the videos, and until we paid up to \$3 per video, no worker picked up our task. (For comparison, to annotate a video~\cite{anonymous}: 3 workers $\times$ 157 questions $\times$ 1 second per question $\times$ \$8/h salary = \$1.) To reduce the base cost to a more manageable \$1 per video, we have used the following strategies:

\noindent {\bf Worker Recruitment.} To overcome the inconvenience threshold, worker recruitment was increased through sign-up bonuses ($211\%$ increased new worker rate) where we awarded a \$5 bonus for the first submission. This increased the total cost by $17\%$. In addition, ``recruit a friend'' bonuses (\$5 if a friend submits 15 videos) were introduced, and were claimed by $4\%$ of the workforce, generating indeterminate outreach to the community. US, Canada, UK, and, for a time, India were included in this study. The first three accounted for estimated $73\%$ of the videos, and $59\%$ of the peak collection rate.

\noindent {\bf Worker Retention.} Worker retention was mitigated through performance bonuses every 15th video, and while only accounting for a $33\%$ increase in base cost, significantly increased retention ($34\%$ increase in come-back workers), and performance ($109\%$ increase in output per worker). 

Each submission in this phase was manually verified by other workers to enforce quality control, where a worker was required to select the corresponding sentence from a line-up after watching the video. The rate of collection peaked at $1225$ per day from $72$ workers. The final cost distribution was: $65\%$ base cost per video, $21\%$ performance bonuses, $11\%$ recruitment bonuses, and $3\%$ verification. The code and interfaces will be made publicly available along with the dataset.

\subsection{Annotations}
Using the generated scripts, all (verb,proposition,noun) triplets were analyzed, and the most frequent grouped into 157 action classes (\eg, \textit{pouring into cup}, \textit{running}, \textit{folding towel}, etc.). The distribution of those is presented in Figure~\ref{fig:mega}.

For each recorded video we have asked other workers to watch the video and describe what they have observed with a sentence (this will be referred to as a \textit{description} in contrast to the previous \textit{script} used to generate the video). We use the original script and video descriptions to automatically generate a list of interacted objects for each video. Such lists were verified by the workers. Given the list of (verified) objects, for each video we have made a short list of 4-5 actions (out of 157) involving corresponding object interactions and asked the workers to verify the presence of these actions in the video.

In addition, to minimize the number of missing labels, we expanded the annotation procedure to exhaustively annotate all actions in the video using state-of-the-art crowdsourcing practices~\cite{anonymous}, where we focused particularly on the test set.

Finally, for all the chosen action classes in each video, another set of workers was asked to label the starting and ending point of the activity in the video, resulting in a temporal interval of each action. A visualization of the data collection process is illustrated in Figure~\ref{fig:mturkpipeline}. On the website we show numerous additional examples from the dataset with annotated action classes.

\section{Charades v1.0 Analysis}

\begin{figure}[pt]
    \centering
    \includegraphics[width=1.0\linewidth]{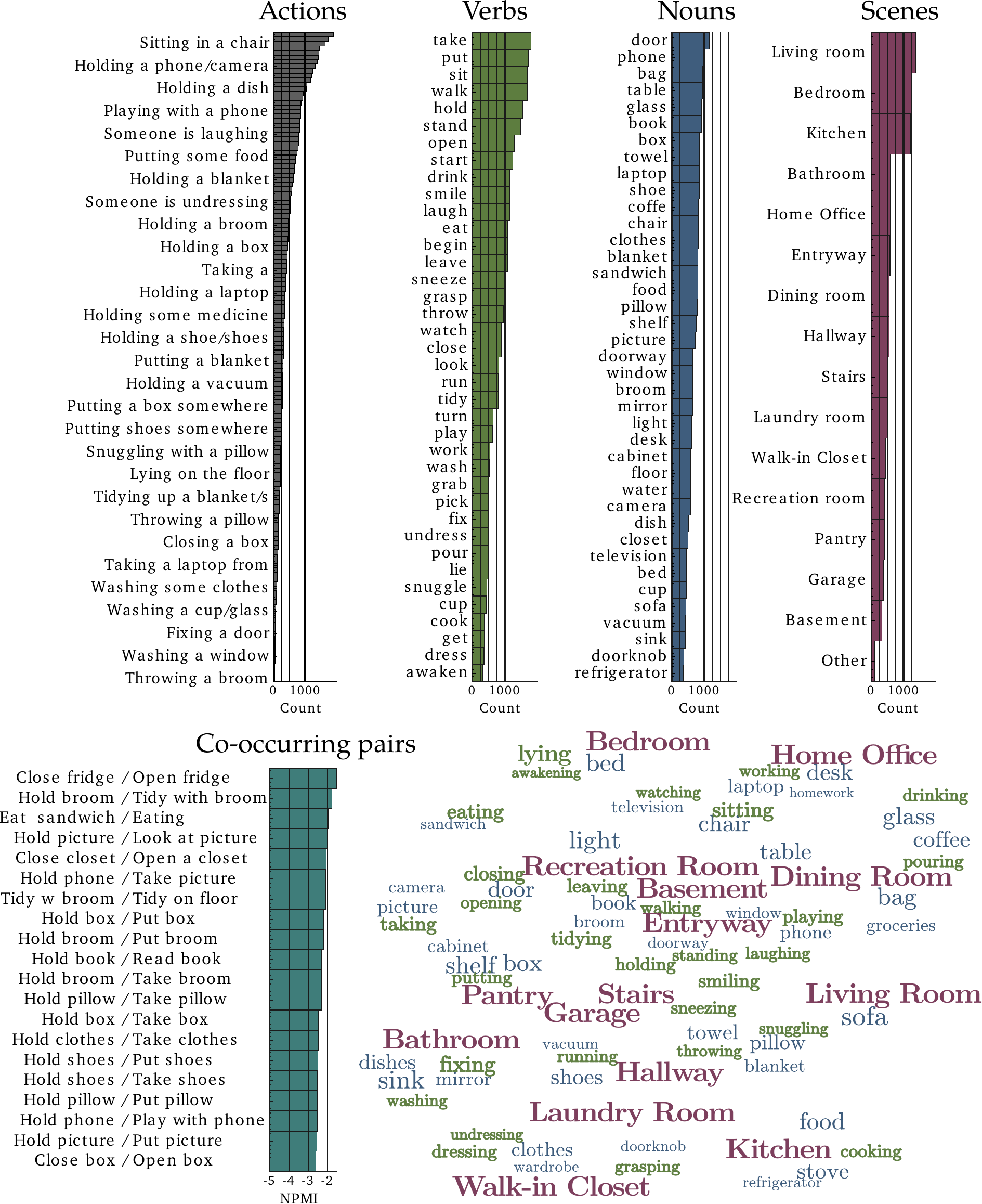}
    \caption{Statistics for actions (gray, every fifth label shown), verbs (green), nouns (blue), scenes (red), and most co-occurring pairs of actions (cyan). Co-occurrence is measured with normalized pointwise mutual information. In addition, a T-SNE embedding of the co-occurrence matrix is presented. We can see that while there are some words that strongly associate with each other (\eg, lying and bed), many of the objects and actions co-occur with many of the scenes. (Action names are abbreviated as necessary to fit space constraints.)}
    \label{fig:mega}
\end{figure}

\emph{Charades} is built up by combining 40 objects and 30 actions in 15 scenes. This relatively small vocabulary, combined with open-ended writing, creates a dataset that has substantial coverage of a useful domain. Furthermore, these combinations naturally form action classes that allow for standard benchmarking. In Figure~\ref{fig:mega} the distributions of action classes, and most common nouns/verbs/scenes in the dataset are presented. The natural world generally follows a long-tailed distribution~\cite{zipf1935psycho,simoncelli2001natural}, but we can see that the distribution of words in the dataset is relatively even. In Figure~\ref{fig:mega} we also present a visualization of what scenes, objects, and actions occur together. By embedding the words based on their co-occurance with other words using T-SNE~\cite{van2008visualizing}, we can get an idea of what words group together in the videos of the dataset, and it is clear that the dataset possesses real-world intuition. For example, \emph{food}, and \emph{cooking} are close to \emph{Kitchen}, but note that except for \emph{Kitchen}, \emph{Home Office}, and \emph{Bathroom}, the scene is not highly discriminative of the action, which reflects common daily activities.

\begin{figure}[tbh]
    \centering
    \includegraphics[width=0.8\linewidth]{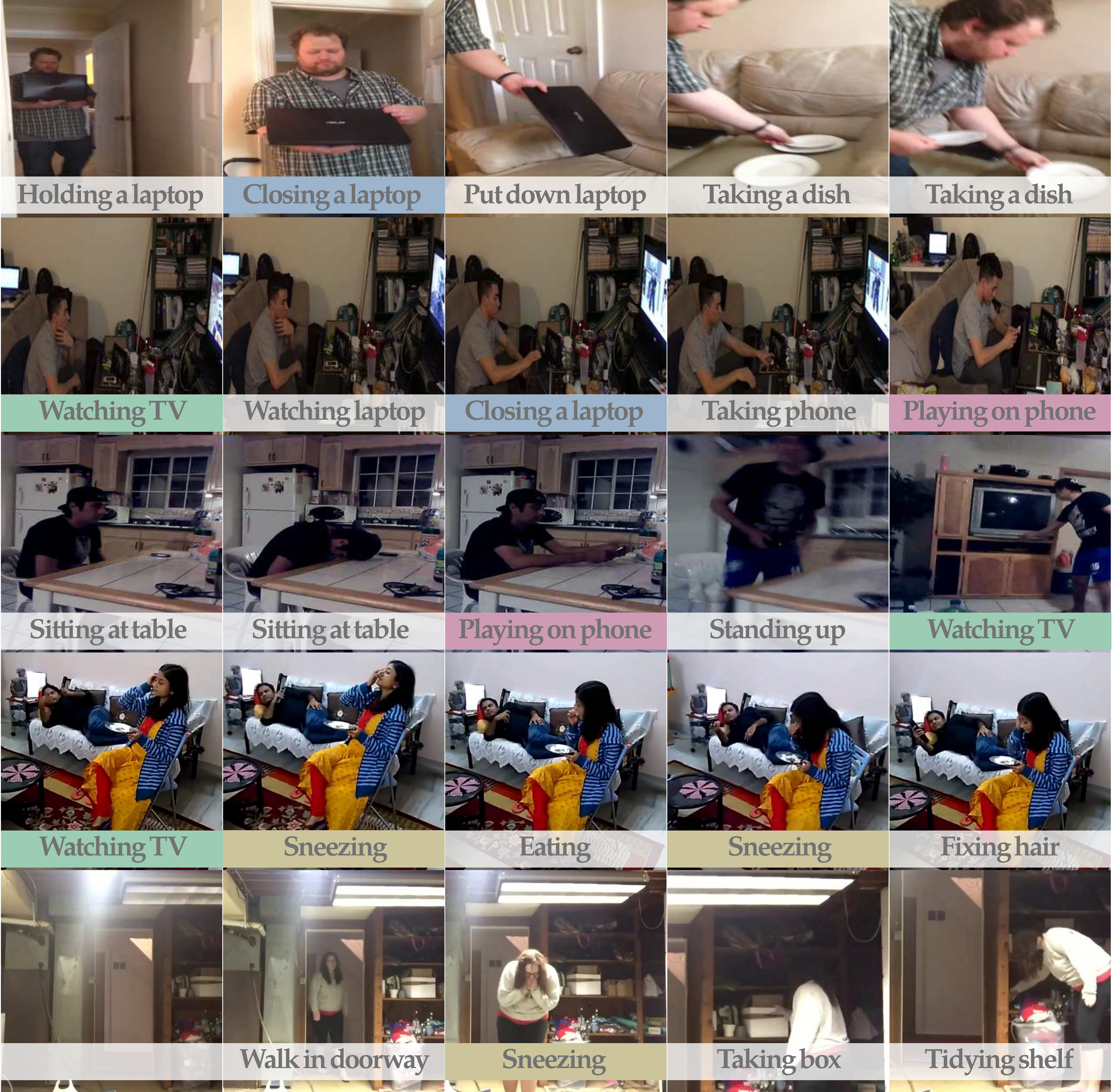}
    \caption{Keyframes from five videos in \textit{Charades}. We see that actions occur together in many different configurations. (Shared actions are highlighed in color).}
    \label{fig:sharedactions}
\end{figure}

\begin{figure}[tbh]
    \centering
    \includegraphics[width=1.0\linewidth]{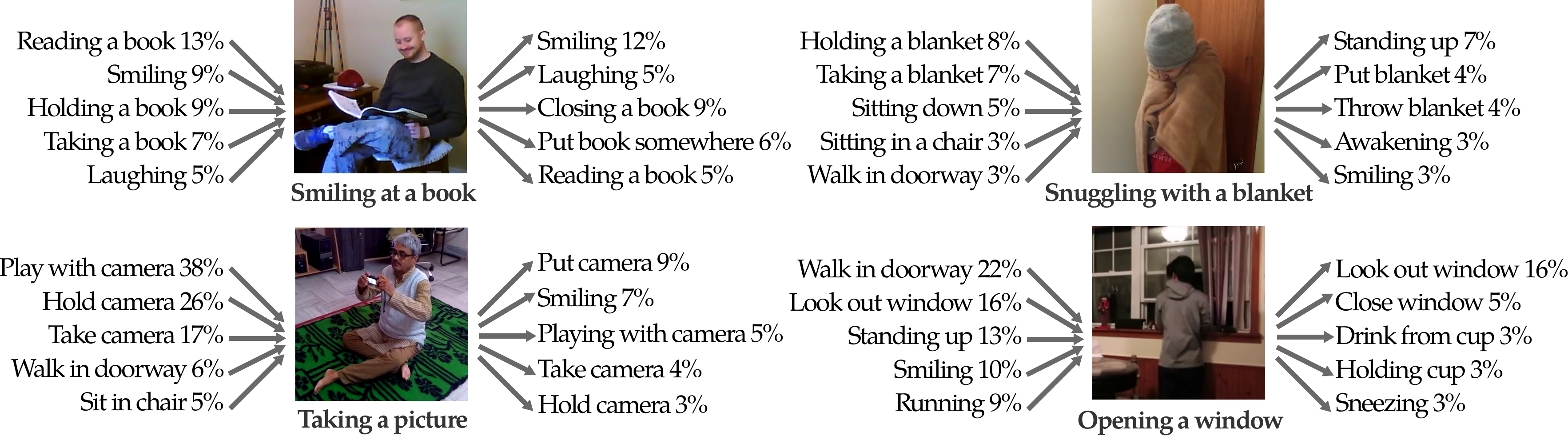}
    \caption{Selected actions from the dataset, along with the top five most probable actions before, and after the action. For example, when \emph{Opening a window}, it is likely that someone was \emph{Standing up} before that, and after opening, \emph{Looking out the window}.}
    \label{fig:beforafter}
\end{figure}

Since we have control over the data acquisition process, instead of using Internet search, there are on average $6.8$ relevant actions in each video. We hope that this may inspire new and interesting algorithms that try to capture this kind of context in the domain of action recognition. Some of the most common pairs of actions measured in terms of normalized pointwise mutual information (NPMI), are also presented in Figure~\ref{fig:mega}.
These actions occur in various orders and context, similar to our daily lives. For example, in Figure~\ref{fig:sharedactions} we can see that among these five videos, there are multiple actions occurring, and some are in common.
We further explore this in Figure~\ref{fig:beforafter}, where for a few actions, we visualize the most probable actions to precede, and most probable actions to follow that action. As the scripts for the videos are generated by people imagining a boring realistic scenario, we find that these statistics reflect human behaviour.

\section{Applications}
\label{sec:applications}

We run several state-of-the-art algorithms on Charades to provide the community with a benchmark for recognizing human activities in realistic home environments. Furthermore, the performance and failures of tested algorithms provide insights into the dataset and its properties.

\noindent \textbf{Train/test set.}
For evaluating algorithms we split the dataset into train and test sets by considering several constraints: (a) the same worker should not appear in both training and test; (b) the distribution of categories over the test set should be similar to the one over the training set; (c) there should be at least 6 test videos and 25 training videos in each category; (d) the test set should not be dominated by a single worker. We randomly split the workers into two groups (80\% in training) such that these constraints were satisfied. The resulting training and test sets contain $7{,}985$ and $1{,}863$ videos, respectively. The number of annotated action intervals are $49{,}809$ and $16{,}691$ for training and test.

\subsection{Action Classification}

Given a video, we would like to identify whether it contains one or several actions out of our 157 action classes. 
We evaluate the classification performance for several baseline methods. Action classification performance is evaluated with the standard mean average precision (mAP) measure. A single video is assigned to multiple classes and the distribution of classes over the test set is not uniform. The label precision for the data is $95.6\%$, measured using an additional verification step, as well as comparing against a ground truth made from 19 iterations of annotations on a subset of 50 videos. We now describe the baselines.

\noindent \textbf{Improved trajectories.} We compute improved dense trajectory features (IDT) \cite{wang2013idt} capturing local shape and motion information with MBH, HOG and HOF video descriptors. We reduce the dimensionality of each descriptor by half with PCA, and learn a separate feature vocabulary for each descriptor with GMMs of 256 components. Finally, we encode the distribution of local descriptors over the video with Fisher vectors~\cite{perronnin2010fv}. A one-versus-rest linear SVM is used for classification. Training on untrimmed intervals gave the best performance.

\begin{table}[t]
\centering
\caption{mAP (\%) for action classification with various baselines.}
\label{tbl:classification}
\setlength{\tabcolsep}{8pt}
\resizebox{\textwidth}{!}{%
\begin{tabular}{ccccccc} \toprule
Random & C3D & AlexNet  & Two-Stream-B & Two-Stream & IDT  & Combined  \\ \hline \rule{0pt}{3ex}
5.9 & 10.9 & 11.3 & 11.9 & 14.3 & 17.2 & 18.6 \\
\bottomrule
\end{tabular}
}
\end{table}
\begin{table}[t]
\centering
\caption{Action classification evaluation with the state-of-the-art approach on Charades. We study different parameters for improved trajectories, by reporting for different local descriptor sets and different number of GMM clusters. Overall performance improves by combining all descriptors and using a larger descriptor vocabulary.}
\label{tbl:classification_idt}
\setlength{\tabcolsep}{7pt}
\begin{tabular}{lccccc} \toprule
        & HOG  & HOF   & MBH   & HOG+MBH  & HOG+HOF+MBH  \\ \hline \rule{0pt}{3ex}
K=64   & 12.3 & 13.9 & 15.0 & 15.8    & 16.5        \\
K=128  & 12.7 & 14.3 & 15.4 & 16.2    & 16.9        \\
K=256  & 13.0 & 14.4 & 15.5 & 16.5    & 17.2         \\
\bottomrule
\end{tabular}
\end{table}

\noindent \textbf{Static CNN features.} In order to utilize information about objects in the scene, we make use of deep neural networks pretrained on a large collection of object images. We experiment with VGG-16~\cite{simonyan2015vgg} and AlexNet~\cite{alexnet2012} to compute $\mathrm{fc}_6$ features over 30 equidistant frames in the video. These features are averaged across frames, L2-normalized and classified with a one-versus-rest linear SVM.
Training on untrimmed intervals gave the best performance.

\noindent \textbf{Two-stream networks.} We use the VGG-16 model architecture~\cite{Simonyan14c} for both networks and follow the training procedure introduced in Simonyan et al.~\cite{simonyan2014twostream}, with small modifications. For the spatial network, we applied finetuning on ImageNet pre-trained networks with different dropout rates. The best performance was with $0.5$ dropout rate and finetuning on all fully connected layers. The temporal network was first pre-trained on the UCF101 dataset and then similarly finetuned on conv4, conv5, and fc layers. Training on trimmed intervals gave the best performance.

\noindent \textbf{Balanced two-stream networks.} We adapt the previous baseline to handle class imbalance. We balanced the number of training samples through sampling, and ensured each minibatch of 256 had at least 50 unique classes (each selected uniformly at random). Training on trimmed intervals gave the best performance.

\noindent \textbf{C3D features.} Following the recent approach from~\cite{tran2015c3d}, we extract $\mathrm{fc}_6$ features from a 3D convnet pretrained on the Sports-1M video dataset~\cite{karpathy2014large}. These features capture complex hierarchies of spatio-temporal patterns given an RGB clip of 16 frames. Similar to~\cite{tran2015c3d}, we compute features on chunks of 16 frames by sliding 8 frames, average across chunks, and use a one-versus-rest linear SVM.
Training on untrimmed intervals gave the best performance.

\noindent Action classification results are presented in Table~\ref{tbl:classification}, where we additionally consider \textbf{Combined} which combines all the other methods with late fusion. 

\begin{figure}[t]
    \centering
    \includegraphics[width=0.8\linewidth]{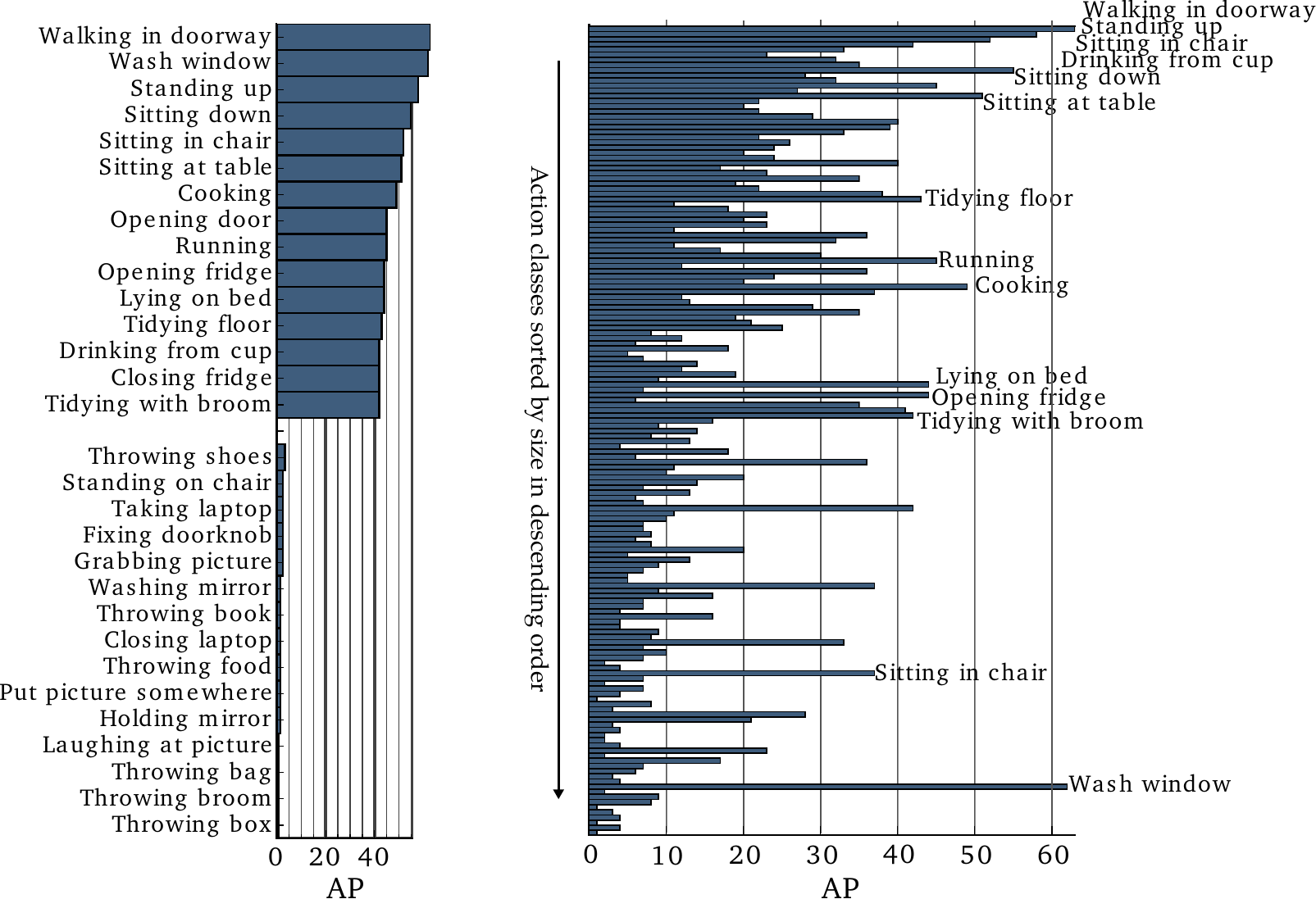}
    \caption{On the left classification accuracy for the 15 highest and lowest actions is presented for \emph{Combined}. On the right, the classes are sorted by their size. The top actions on the left are annotated on the right. We can see that while there is a slight trend for smaller classes to have lower accuracy, many classes do not follow that trend.}
    \label{fig:classification_analysis}
\end{figure}

Notably, the accuracy of the tested state-of-the-art baselines is much lower than in most currently available benchmarks. Consistently with several other datasets, IDT features~\cite{wang2013idt} outperform other methods by obtaining $17.2\%$ mAP. To analyze these results, Figure~\ref{fig:classification_analysis}(left) illustrates the results for subsets of best and worst recognized action classes. We can see that while the mAP is low, there are certain classes that have reasonable performance, for example \emph{Washing a window} has $62.1\%$ AP.
\begin{figure}[t]
    \centering
    \includegraphics[width=0.8\linewidth]{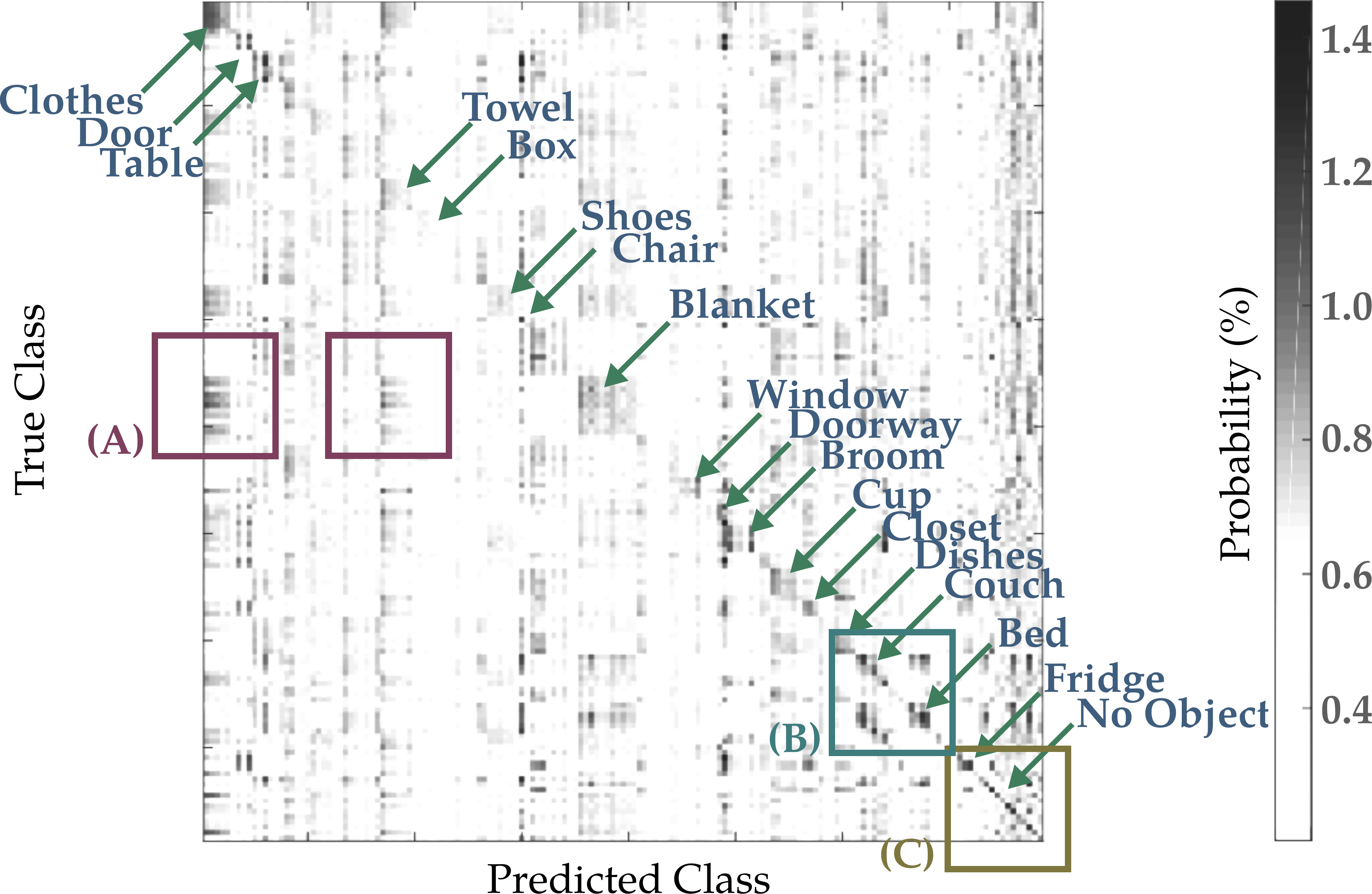}
    \caption{Confusion matrix for the \emph{Combined} baseline on the classification task. Actions are grouped by the object being interacted with. Most of the confusion is with other actions involving the same object (squares on the diagonal), and we highlight some prominent objects. Note: (A) High confusion between actions using \emph{Blanket}, \emph{Clothes}, and \emph{Towel}; (B) High confusion between actions using \emph{Couch} and \emph{Bed}; (C) Little confusion among actions with no specific object of interaction (\eg \emph{standing up}, \emph{sneezing}). }
    \label{fig:confusion}
\end{figure}
To understand the source of difference in performance for different classes, Figure~\ref{fig:classification_analysis}(right) illustrates AP for each action, sorted by the number of examples, together with names for the best performing classes. 
The number of actions in a class is primarily decided by the universality of the action (can it happen in any scene), and if it is common in typical households (writer bias).
It is interesting to notice, that while there is a trend for actions with higher number of examples to have higher AP, it is not true in general, and actions such as \emph{Sitting in chair}, and \emph{Washing windows} have top-15 performance.

Delving even further, we investigate the confusion matrix for the \emph{Combined} baseline in Figure~\ref{fig:confusion}, where we convert the predictor scores to probabilities and accumulate them for each class. For clearer analysis, the classes are sorted by the object being interacted with. The first aspect to notice is the squares on the diagonal, which imply that the majority of the confusion is among actions that interact with the same object (\eg, \emph{Putting on clothes}, or \emph{Taking clothes from somewhere}), and moreover, there is confusion among objects with similar functional properties. The most prominent squares are annotated with the object being shared among those actions. The figure caption contains additional observations. While there are some categories that show no clear trend, we can observe less confusion for many actions that have no specific object of interaction. Evaluation of action recognition on this subset results in $38.9\%$ mAP, which is significantly higher than average. Recognition of fine-grained actions involving interactions with the same object class appears particularly difficult even for the best methods available today. We hope our dataset will encourage new methods addressing activity recognition for complex person-object interactions.

\subsection{Sentence Prediction}

Our final, and arguably most challenging task, concerns prediction of free-from sentences describing the video. Notably, our dataset contains sentences that have been used to create the video (\emph{scripts}), as well as multiple video \emph{descriptions} obtained manually for recorded videos. The scripts used to create videos are biased by the vocabulary, and due to the writer's imagination, generally describe different aspects of the video than descriptions. The description of the video by other people is generally simpler and to the point. Captions are evaluated using the CIDEr, BLEU, ROUGE, and METEOR metrics, as implemented in the COCO Caption Dataset~\cite{capeval2015}. These metrics are common for comparing machine translations to ground truth, and have varying degrees of similarity with human judgement. For comparison, human performance is presented along with the baselines where workers were similarly asked to watch the video and describe what they observed.
We now describe the sentence prediction baselines in detail:

\begin{table}[t]
\centering
\caption{Sentence Prediction. In the \emph{script} task one sentence is used as ground truth, and in the \emph{description} task $2.4$ sentences are used as ground truth on average. We find that S2VT is the strongest baseline.}
\label{tbl:caption}
\begin{tabular}{r@{\hskip .15in}cccccl@{\hskip .1in}ccccc} \toprule
                                      & \multicolumn{5}{c}{\textit{\textbf{Script}}} &  & \multicolumn{5}{c}{\textit{\textbf{Description}}} \\
                                      & RW       & Random    & NN      & S2VT    & Human    &  & RW      & Random    & NN      & S2VT    & Human   \\ \cline{2-6} \cline{8-12}
\rule{0pt}{3ex} CIDEr            & 0.03     & 0.08      & 0.11    & 0.17    & 0.51     &  & 0.04    & 0.05      & 0.07    & 0.14    & 0.53    \\
$\text{BLEU}_4$  & 0.00     & 0.03      & 0.03    & 0.06    & 0.10     &  & 0.00    & 0.04      & 0.05    & 0.11    & 0.20    \\
$\text{BLEU}_3$  & 0.01     & 0.07      & 0.07    & 0.12    & 0.16     &  & 0.02    & 0.09      & 0.10    & 0.18    & 0.29    \\
$\text{BLEU}_2$  & 0.09     & 0.15      & 0.15    & 0.21    & 0.27     &  & 0.09    & 0.20      & 0.21    & 0.30    & 0.43    \\
$\text{BLEU}_1$  & 0.37     & 0.29      & 0.29    & 0.36    & 0.43     &  & 0.38    & 0.40      & 0.40    & 0.49    & 0.62    \\
$\text{ROUGE}_L$ & 0.21     & 0.24      & 0.25    & 0.31    & 0.35     &  & 0.22    & 0.27      & 0.28    & 0.35    & 0.44    \\
METEOR           & 0.10     & 0.11      & 0.12    & 0.13    & 0.20     &  & 0.11    & 0.13      & 0.14    & 0.16    & 0.24    \\ 
\bottomrule
\end{tabular}
\end{table}

\begin{figure}
    \centering
    \includegraphics[width=0.8\linewidth]{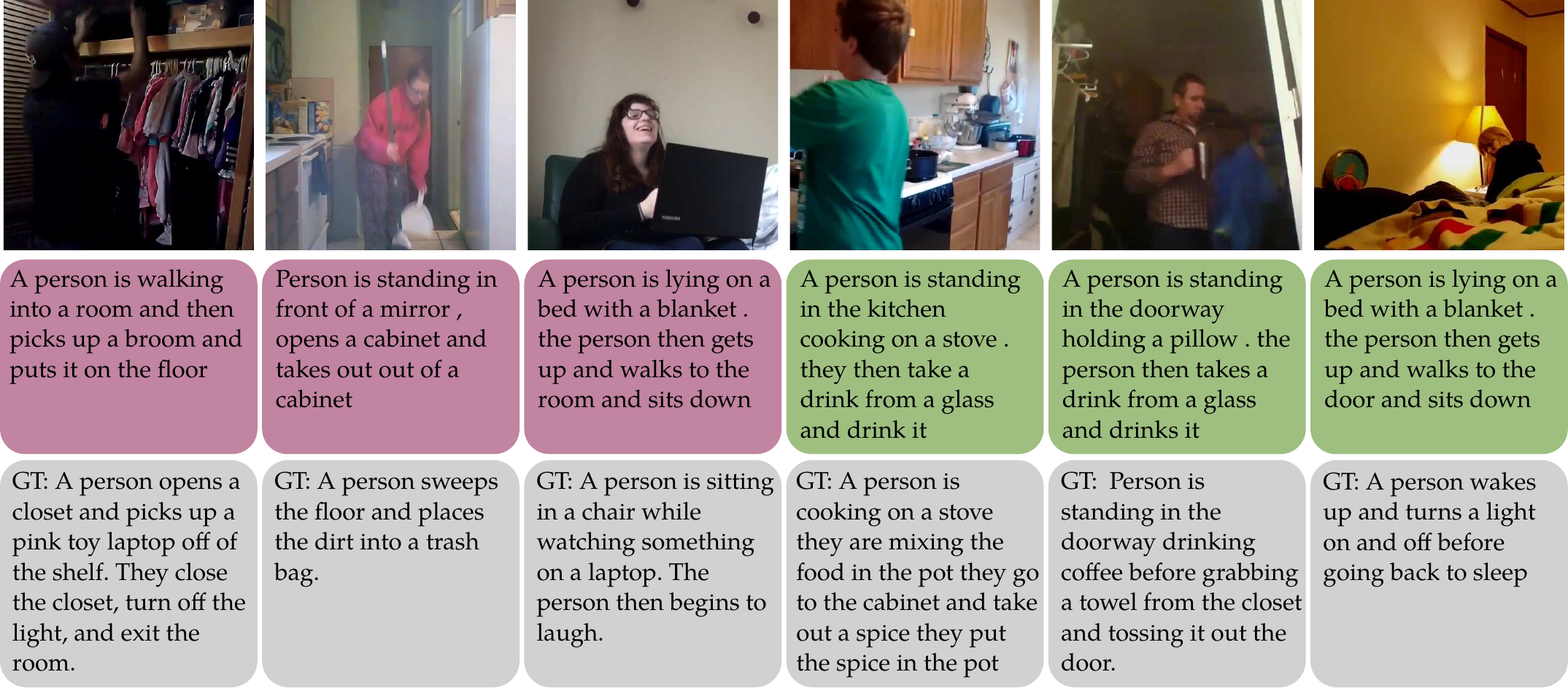}
    \caption{Three generated captions that scored low on the CIDEr metric (red), and three that scored high (green) from the strongest baseline (S2VT). We can see that while the captions are fairly coherent, the captions lack sufficient relevance.}
    \label{fig:captionresults}
\end{figure}

\noindent \textbf{Random Words (RW):} Random words from the training set. 

\noindent \textbf{Random Sentence (Random):} Random sentence from the training set.

\noindent \textbf{Nearest Neighbor (NN):} Inspired by Devlin et al.~\cite{devlin2015exploring} we simply use a 1-Nearest Neighbor baseline computed using AlexNet $\mathrm{fc}_7$ outputs averaged over frames, and use the caption from that nearest neighbor in the training set.

\noindent \textbf{S2VT:} We use the S2VT method from Venugopalan et al.~\cite{venugopalan2015sequence}, which is a combination of a CNN, and a LSTM.

Table~\ref{tbl:caption} presents the performance of multiple baselines on the caption generation task. We both evaluate on predicting the \emph{script}, as well as predicting the \emph{description}.
As expected, we can observe that descriptions made by people after watching the video are more similar to other descriptions, rather than the scripts used to generate the video.
Table~\ref{tbl:caption} also provides insight into the different evaluation metrics, and it is clear that CIDEr offers the highest resolution, and most similarity with human judgement on this task. In Figure~\ref{fig:captionresults} few examples are presented for the highest scoring baseline (S2VT). We can see that while the language model is accurate (the sentences are coherent), the model struggles with providing relevant captions, and tends to slightly overfit to frequent patterns in the data (\eg, \emph{drinking from a glass/cup}).

\section{Conclusions}

We proposed a new approach for building datasets. Our Hollywood in Homes approach allows not only the labeling, but the data gathering process to be crowdsourced. In addition, \emph{Charades} offers a novel large-scale dataset with diversity and relevance to the real world. We hope that Charades and Hollywood in Homes will have the following benefits for our community:

\noindent \emph{(1) Training data}: Charades provides a large-scale set of $66{,}500$ annotations of actions with unique realism.

\noindent \emph{(2) A benchmark}: Our publicly available dataset and provided baselines enable benchmarking future algorithms.

\noindent \emph{(3) Object-action interactions}: The dataset contains significant and intricate object-action relationships which we hope will inspire the development of novel computer vision techniques targeting these settings.

\noindent \emph{(4) A framework to explore novel domains}: We hope that many novel datasets in new domains can be collected using the Hollywood in Homes approach.

\noindent \emph{(5) Understanding daily activities}: Charades provides data from a unique human-generated angle, and has unique attributes, such as complex co-occurrences of activities. This kind of realistic bias, may provide new insights that aid robots equipped with our computer vision models operating in the real world.

\section{Acknowledgements}

This work was partly supported by ONR MURI N00014-16-1-2007, ONR N00014-13-1-0720, NSF IIS-1338054, ERC award ACTIVIA, Allen Distinguished Investigator Award, gifts from Google, and the Allen Institute for Artificial Intelligence. The authors would like to thank: Nick Rhinehart and the anonymous reviewers for helpful feedback on the manuscript; Ishan Misra for helping in the initial experiments; and Olga Russakovsky, Mikel Rodriguez, and Rahul Sukhantakar for invaluable suggestions and advice. Finally, the authors want to extend thanks to all the workers at Amazon Mechanical Turk.

\bibliographystyle{splncs}
\bibliography{eccv2016gunnar}

\end{document}